\documentclass{ametsocV6.1}
\usepackage{subcaption}
\usepackage{float}
\usepackage{amsmath}

\title{Capturing Climatic Variability: Using Deep Learning for Stochastic Downscaling}

\authors{Kiri Daust\aff{a,b}\correspondingauthor{Kiri Daust, kiridaust@uvic.ca}, Adam Monahan\aff{a}}

\affiliation{\aff{a}{University of Victoria}\\
\aff{b}{Future Forest Ecosystem Centre, BC Ministry of Forests}
}

\abstract{Adapting to the changing climate requires accurate local climate information, a computationally challenging problem. Recent studies have used Generative Adversarial Networks (GANs), a type of deep learning, to learn complex distributions and downscale climate variables efficiently. Capturing variability while downscaling is crucial for estimating uncertainty and characterising extreme events—critical information for climate adaptation. Since downscaling is an undetermined problem, many fine-scale states are physically consistent with the coarse-resolution state. To quantify this ill-posed problem, downscaling techniques should be stochastic, able to sample realisations from a high-resolution distribution conditioned on low-resolution input. Previous stochastic downscaling attempts have found substantial underdispersion, with models failing to represent the full distribution. We propose approaches to improve the stochastic calibration of GANs in three ways: a) injecting noise inside the network, b) adjusting the training process to explicitly account for the stochasticity, and c) using a probabilistic loss metric. We tested our models first on a synthetic dataset with known distributional properties, and then on a realistic downscaling scenario, predicting high-resolution wind components from low-resolution climate covariates. Injecting noise, on its own, substantially improved the quality of conditional and full distributions in tests with synthetic data, but performed less well for wind field downscaling, where models remained underdispersed. For wind downscaling, we found that adjusting the training method and including the probabilistic loss improved calibration. The best model, with all three changes, showed much improved skill at capturing the full variability of the high-resolution distribution and thus at characterising extremes.}

\begin{document}

\maketitle

\statement
Estimating extremes at local spatial scales is crucial for climate adaptation. Most approaches for increasing the resolution of climate data focus on mean values; other methods are too computationally intensive for wide-scale use. We present a computationally efficient method for increasing the resolution of climate variables that improves estimates of extreme events and thus assists with planning in socio-ecological systems.

\section{Introduction}
As society tries to adapt to Earth's changing climate, access to accurate, local-scale climate information is essential. Earth System Models (ESMs) provide state-of-the-art projections on a global scale, but provide insufficient spatial resolution for regional analyses. Thus, downscaling - creating high-resolution climate information from low-resolution, large-scale data - is an important practical tool. Spatially and temporally high resolution fields of climate data over large scales are important for many applications, including simulation of extreme events at local scales \citep[e.g. fires, floods, storms;][]{fischer2021increasing}, local planning, and making climate informed ecological decisions such as tree species selection \citep{mackenzie2021ecological}. Hence, accurate and computationally efficient downscaling methods are crucial for local climate adaptation. 

Strategies for downscaling coarse-resolution climate model simulations to regional scales can be broadly classified into dynamic downscaling and statistical downscaling. Dynamic downscaling employs a limited-area numerical climate model to resolve fine-scale features,  driven by large-scale weather patterns from the low resolution ESM \citep{skamarock2001prototypes}. Dynamic downscaling can capture complex spatial patterns at smaller scales, and can provide high-accuracy downscaling. However, it is highly computationally intensive, and can only be used for relatively short time periods. Statistical downscaling develops statistical relationships between low-resolution (LR) and high-resolution (HR) climate variables. Statistical downscaling can be applied either at individual points, often using a combination of bias correction and transformation of statistical moments \citep{maraun2013bias}, or can be used to downscale entire fields. Common techniques for the latter include parametric approaches (e.g., Gaussian Process/kriging), where covariances are specified to allow analytic solutions. While these approaches have shown some success, climate-field downscaling methods make strong assumptions about the distribution and homogeneity of statistics, which are often not satisfied \citep{chen2014spatial}. Also, many current methodologies struggle to accurately downscale spatially complex  variables (i.e., variables with non-linear dependence on elevation or spatially heterogeneous dependence structures) and capture extremes \citep{harris2022generative}. Recently, a new strategy for statistical downscaling over climate fields has been developed that uses deep (i.e., many layered) learning algorithms to learn a mapping from LR to  HR paired climate fields \citep{annau2023algorithmic}.  It has been found that this strategy can produce downscaled fields with much higher accuracy than traditional statistical downscaling, and does not require the prohibitive computation required for dynamic downscaling. 

Downscaling is intrinsically an underdetermined problem with a distribution of possible HR realisations physically consistent with any given LR input \citep{afshari2023statistical}. This is especially true since weather is sensitive to initial conditions: minute differences can result in drastically different outcomes due to development of internal variability \citep{lucas2021convection}. Stochastic Weather Generators, which attempt to sample from the distribution of weather states, have been used to try and account for this variability \citep{wilks2010use}. Being able to capture the full variability of a downscaling problem is crucial for quantifying uncertainty, and for characterising extreme events. Ideally, downscaling techniques would allow sampling from the HR distribution, conditioned on the LR input. 

Generative Adversarial Networks (GANs) have been successful in various generative AI fields, especially computer vision applications \citep{goodfellow2014generative}. GANs use two separate deep convolutional networks during training: a Generator network which is given input and attempts to create plausible counterfeits of the training data; and a discriminator or Critic network which is provided training data mixed with Generator output and attempts to distinguish between the counterfeits and the real data. During training, the two networks play a minimax game: the Generator tries to improve its output to ``fool" the discriminator and the discriminator tries to improve its ability at distinguishing between real and generated samples. In the last few years, GANs have been introduced to deep-learning-based downscaling and have shown success in drawing realizations from high-dimensional non-Gaussian distributions with complicated dependence structures . Conditional GANs, developed by \citet{mirza2014conditional} allow the GAN to draw realisations from distributions, conditioned on covariates.  

Much of the development of GANs for climate downscaling builds on work from the computer vision field of super resolution. Most studies in computer vision use conditional GANs, where the networks are provided LR  information and learn to sample from the HR conditional distribution \citep{ledig2017photo}. With the introduction of GANs to climate downscaling, difficulties with instability during training \citep{wang2020state} were improved by the introduction of the Wasserstein GAN \citep[WGAN;][]{arjovsky2017wasserstein}. In the WGAN, instead of the Critic network estimating the probability of individual realisations being real, it estimates the Wasserstein distance between the true HR distribution and the generated distribution. Not only does this substantially improve stability during training, but for downscaling, it conceptually makes sense to focus on convergence in distribution of generated and truth fields. 

The initial formulation of (unconditional) GANs used a stochastic approach where the only input to the Generator was Gaussian noise, generating different realisations for each different noise input. With the development of conditional GANs \citep{mirza2014conditional, ledig2017photo}, and the subsequent Super Resolution GAN (SRGAN) and Enhanced Super Resolution GAN (ESRGAN) frameworks from the field of super resolution, the noise input was replaced by the conditioning fields, leading to a semi-deterministic network. In this setting, each trained Generator would still draw a realisation from the conditional distribution, but it would always draw the same realisation for each set of conditioning fields (theoretically, one could draw a different realisation by training a new model). To allow for explicit stochasticity in conditional GANs, recent studies have used variations of adding noise covariates, stacked with the LR conditioning fields. \citet{price2022increasing} concatenated a noise layer part way through their Generator network, while \citet{harris2022generative} concatenated multiple noise inputs with the conditioning information at the beginning of the network. Both studies found that the stochastic results were underdispersive: trained models were unable to capture the full range of variability, often only sampling from the centre of the conditional distribution. Recent advancements in Super Resolution \citep[e.g. nESRGAN+,][]{nesrganplus} have improved stochastic calibration, but have not yet adapted it to climate downscaling. Furthermore, most downscaling studies using stochastic GANs have focused their analyses on image quality; to our knowledge, no research in climate downscaling has fully investigated the ability of stochastic GANs to learn and sample from the conditional HR distribution.   

We aim to fill this gap by improving stochastic GAN frameworks for climate downscaling. Most of this work builds on \citet{annau2023algorithmic}. While their model showed success for downscaling wind fields, it was not fully stochastic. We use a similar model architecture as \citet{annau2023algorithmic} adapted for full stochasticity. An obvious challenge with testing distributional quality in a downscaling setting is the lack of a truth conditional distribution, as in most applications we only have access to a single truth realisation for each timestep. To address this challenge, we first consider an idealised approach based on synthetic data with known distributional properties. Based on these experiments, we test a ``noise injection" method, where hundreds of noise fields, at different spatial resolutions, are injected into the latent layers of the network. This approach provides excellent stochastic calibration on the synthetic data. We then test our modification on a real-world downscaling problem, predicting HR wind components from LR conditioning data. Challenges with underdisperison on the wind data lead to development of an updated loss function using a probabilistic error function and modification of the training method to fully utilise the stochasticity. Our final model is successful at capturing variability, and improves estimates of moderate extremes. 

\section{Methods}
All models in this paper use the same basic super-resolution structure. We train the models on paired sets of LR conditioning fields (covariates) and HR truth fields. The GAN then learns a mapping to the HR fields from the input covariates. For consistency, we keep the same resolution and size of fields across all models: HR fields are 128x128 pixels, and LR fields are 16x16 pixels, resulting in a downscaling factor of eight. 

\subsection{Data}

\subsubsection{Synthetic Data}
Evaluating the distributional quality of stochastic realisations on realistic downscaling problems is challenging, as there is rarely more than one realisation of the ``ground truth" to compare with for a given sample of the conditioning field. While certain metrics (e.g. Cumulative Ranked Probability Score (CRPS) or Rank Histogram, discussed below) provide information on model calibration, they do not allow direct comparison of conditional distributions. We thus created a simple synthetic dataset with known distribution properties. To make the HR fields, we added a mean zero Gaussian field with a specified covariance structure to a specified non-stationary mean (exponential in one axis and sigmoidal in the other). This sum was then squared to generate a field with pointwise chi-square marginal distributions. That is, we drew realisations from $r_{ij}$ where
\begin{align}
Y&\sim N(\vec{1},\Sigma) \\
s_{ij}&=\frac{5e^{x_{i}}}{1 + e^{-8y_{j}}} + Y_{ij}\\
r_{ij}&=s_{ij}^2
\end{align}
where $r$ is the output HR field, $x$ and $y$ are the axis values, and $\Sigma$ is a 128 x 128 covariance matrix with correlations decreasing linearly to zero over four pixels along both directions. To ensure large scale structure varied between samples, we randomly scaled and reflected the $x$ and $y$ axis, as follows: for each field, we drew realisation of random variables $A_1, A_2, B_1, B_2$ from a uniform distribution over {-1,0,1}, such that $A_1 \neq A_2$ and $B_1 \neq B_2$ and then rescaled $x$ and $y$ as
\begin{align}
    x_n &= A_1 + \frac{n(A_2 - A_1)}{128}, n \in \{0,1,...,128\}\\
    y_n &= B_1 + \frac{n(B_2 - B_1)}{128}, n \in \{0,1,...,128\}.
\end{align}
To create the LR input fields, we spatially averaged 8 x 8 regions  of the HR fields. 

GANs commonly struggle to capture multi-modal distributions and tend to converge on the conditional mean, a phenomenon known as mode collapse \citep[e.g.,][]{saatci2017bayesian}. To test the ability of our GANs to learn multi-modal distributions,  we generated a second set of realisations of bimodal fields using a Gaussian Mixture Model where $\delta$ is a Bernoulli random variable:
\begin{align}
    X &\sim N(\vec{5},\Sigma)\\
    Y &\sim N(\vec{1},\Sigma)\\
    \delta &\sim B(n = 1, p = 0.35)\\
    s_{ij} &= \frac{5e^{x_{i}}}{1 + e^{-8y_{j}}} + X_{ij}^{\delta} \cdot Y_{ij}^{1 - \delta} \\
    r_{ij} &= s_{ij}^2.
\end{align}

For all synthetic data experiments, training used 5000 pairs of fields; 2000 additional pairs were reserved for testing. To compare marginal distributions, we created a set of 500 fields with the same large-scale spatial pattern (i.e. same $x$ and $y$ scale and rotation), so that the only difference was the added Gaussian/mixture field. These could then be interpreted as ensembles of truth realisations given the same conditioning field, and were used to test generated pixel-wise marginal distributions. 

To investigate the effects of spatial complexity on the generated fields, we created three further synthetic datasets with three different levels of spatial heterogeneity. We used a field of complex topography from the south-coast of British Columbia, and added the same Gaussian field described above. To vary spatial complexity, we scaled the topography with 3 different weights. That is, 
\begin{align}
    s_{ij} &= wZ_{ij} + Y_{ij}, w \in \{0.1,1,10\}\\
    r_{ij} &= s_{ij}^2
\end{align} where $Z$ is the topography field, scaled to zero mean and unit standard variance.  We used weights of 0.1 (field dominated by original dataset) for low heterogeneity, 1 for moderate heterogeneity, and 10 for high heterogeneity (dominated by added topography). 

\subsubsection{Convection Permitting Regional Model Case Study}
Since this research aims to improve deep-learning based downscaling of climate data, it is important to test results on more realistic settings. Here, we modelled HR zonal ($u$) and meridional ($v$) wind components using LR wind components, pressure, temperature, and HR topography. Our architecture follows that of  \citet{annau2023algorithmic} with the exceptions that: (i) HR topography is included as a covariate in the Generator, and (ii) all covariates are also passed to the Critic. Wind is an important climate variable for various applications, but it is often challenging to model due to having complex mesoscale patterns. We consider a square region covering the coastal mountains, in southwestern Canada ($49^\circ$ to $53^\circ$ N, $122^\circ$ to $126^\circ$ W), as its high degree of topographic complexity represents a realistically challenging downscaling scenario. HR wind fields were obtained from WRF runs produced for the CONUS2 simulation driven by ERA Interim  \citep{li2019high}, which contains hourly data for a 14 year period. LR covariates were from ERA5. While ERA-Interim was used to drive the WRF model at the boundary conditions, both ERA5 and ERA-Interim represent the same realisation of the climate system, so it is reasonable to use ERA5 in the paired LR data. 

This HR and LR pairing represents a practical application of downscaling, where the LR and HR fields are from different models \citep[as in][]{annau2023algorithmic}. Many previous studies in deep-learning based downscaling have used idealised pairings, where the LR fields are created by coarsening the HR fields, resulting in perfectly matched pairing (note that this was the approach we used for the synthetic data experiments). As a consequence of natural internal variability, some of the meteorological features on scales common to both resolutions will differ between the LR and HR fields, so our model has to account for such differences, in addition to downscaling. 

To preprocess the data, we first transformed the WRF fields to the ERA5 projection, and then remapped the HR fields to the specified downscaling factor. We then standardised the data to mean zero and unit variance across time and space, and within each covariate, as it standard in machine learning studies \citep{annau2023algorithmic}. Finally, we selected three apparently unexceptional years (no ENSO or evident seasonal extreme wind events) for training (2003, 2008, and 2013), and two years as an unseen test set (2005 and 2012). Hourly data over three years resulted in 26304 samples; sample number was limited by computational constraints. All models were trained on a single NVIDIA RTX 4090 GPU with 24 Gb VRAM.

\subsection{Model}
This work utilizes conditional GANs \citep{mirza2014conditional}, which have shown success at learning the mapping  between low resolution variables and the desired output variables. Specifically, we use the Wasserstein Conditional GAN formulation \citep{arjovsky2017wasserstein}, where the Critic network learns to estimate the Wasserstein distance between the high-dimensional distributions of the generated and true fields. 

\subsubsection{Architecture}
Most GAN network architectures employ dense convolutional blocks, which have been shown to be excellent at extracting representative features from images. Our network architecture is based on the Enhanced Super-Resolution GAN \citep[ESRGAN;][]{wang2018esrgan} setup, using Residual in Residual Dense Blocks as the main convolutional blocks in the Generator. Specifically, we adapt the architecture employed in \citet{annau2023algorithmic} to allow noise input and multiple covariate streams, as follows. 

\begin{figure}[t]
  \noindent\includegraphics[width=36pc,angle=0]{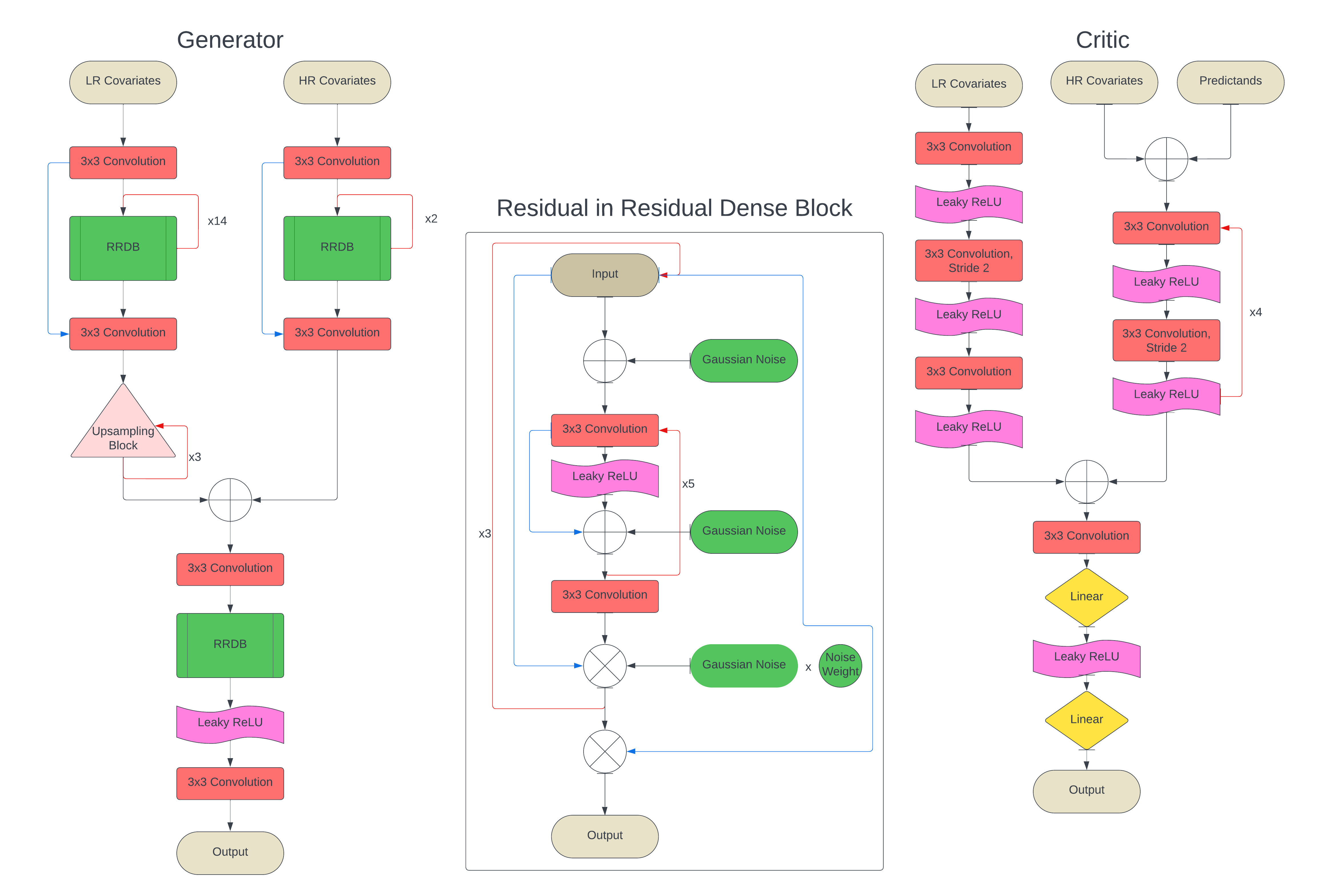}\\
  \caption{Architecture of GAN networks showing Residual in Residual Dense Block (RRDB) with noise injection. Green denotes locations where noise is added into the network. Rectified Linear Units (ReLU) are used to introduce non-linearity.}\label{f0}
\end{figure}

Unlike other applications of super resolution, climate downscaling often has access to  pertinent HR information during training. A common example of such HR information is topography, which influences local climate strongly. While many previous studies have  included topography as a covariate, it has often been input at low resolution with climate covariates, discarding potentially useful information. We therefore created a Generator architecture which allowed us to fully utilise the HR covariates. Using a strategy similar to Depthwise Separable networks \citep{jiang2020single} we created two input streams within the Generator, one stream for each resolution. Each stream has equivalent convolutional blocks applied in parallel, and after the LR stream has passed through the upsampling blocks to increase the resolution, the two streams are concatenated and passed through a final convolutional block (figure \ref{f0}).

In the standard super-resolution formulation, the Critic network is only given samples of the predictands (either generated or from the training data). However, \citet{harris2022generative} found that passing all available covariates to the Critic network can improve its predictions, and in the Wasserstein GAN formulation, it is important that the Critic network is able to make relevant estimates of the Wasserstein distance. Given that we want the Generator to sample from a conditional distribution, we thus want the Critic to make use of conditioning information when quantifying this distribution. Similarly to the Generator, we adapted the Critic network to allow a separate input stream for the LR covariates, which is concatenated to the HR stream after downsampling convolutional blocks (figure \ref{f0}). 

\subsubsection{Noise Injection}
To improve the model's ability to sample across the entire range of the conditional distribution, we adjusted the Generator architecture to inject noise directly into the latent representations produced by the convolutional layers. Specifically, we based our approach on nESRGAN+ \citep{nesrganplus}, and concatenate uncorrelated, mean zero, unit variance Gaussian noise fields with the latent layers inside each Dense Block (figure \ref{f0}). With our architecture, this leads to six noise injection instances in each Residual Dense Block, and 18 noise injections in each Residual in Residual Dense Block (RRDB). Our full noise Generator contains 14 RRDB in the LR input stream, 2 in the HR input stream, and one after concatenation, resulting in 252 LR noise layers and 54 HR noise layers. 

To test the effect of the number of noise injection layers, we replaced some of the stochastic RRDB with deterministic RRDB (i.e., RRDB with no noise injection). Altogether, we tested three different levels of noise injection: low (2 stochastic RRDB-LR, 0 stochastic RRDB-HR), moderate (7 stochastic RRDB-LR, 0 stochastic RRDB-HR), and full noise (14 stochastic RRDB-LR, 1 stochastic RRDB-HR). As a baseline model, we also considered the more standard noise-covariate approach \citep[similar to that used in][]{leinonen2020stochastic}, for which a Gaussian noise field is concatenated with the LR input covariates before passing through the Generator network.  

\subsubsection{Training}
In Wasserstein GAN training, the Critic tries to maximise the difference in the distributional distance between true fields and generated fields, while the Generator attempts to minimize this distance (i.e., it attempts to make it difficult for the Critic to distinguish the generated fields from the training data). Previous studies have found that solely relying on the Critic loss (adversarial loss) for the Generator training can lead to instability, such that the training process does not converge \citep{wang2018esrgan}. Here, we use the common approach of adding an extra content loss term to the Generator - a pixel-wise error metric between the training data and the generated fields, intended to aid convergence at large scales. While standard content loss formulations are deterministic (e.g., Mean Absolute Error) and emphasize convergence in realization, it is also possible to use probabilistic measures (e.g. CRPS) which emphasize convergence in conditional distribution.

Since our goal is to sample realisations from the entire conditional distribution of HR fields, we do not want the loss function to force the Generator to create copies of the training data - we consider the ground truth as one realisation of the conditional distribution. Ideally, we would expect generated realisation to have similar statistics and large-scale features as the training data, but not be identical. Over-reliance on content loss (particularly deterministic measures) can degrade model performance, as it overly penalises small deviations in feature location/presence. In situations where features are spatially shifted, pixel wise error metrics such as the content loss will penalize the model twice: once for the feature not occurring where it does in the ground truth, and once for the feature occurring where it is not in the ground truth. This double-penalty problem is a well-known issue with pixel-based losses in generative networks \citep{harris2022generative}, and results in overly blurry output, where the model converges on the conditional median, thus suppressing small-scale features and extremes \citep{annau2023algorithmic}. While the adversarial loss in GANs (in our case the Wasserstein distance) is not a pixel-wise metric and does not constrain the network in the same way, the use of content losses can suppress variability. 

We considered two training techniques in our models to address the double penalty problem while rewarding convergence at large scales: frequency separation, and stochastic sampling. Frequency separation, introduced by \citet{annau2023algorithmic}, passes all spatial frequencies to the adversarial loss but only the low frequencies to the content loss, allowing the model to more freely develop high frequency patterns. Stochastic sampling is an approach modified from \citet{harris2022generative}, where the conditional average across stochastic realisations is passed to the content loss. In each Generator training step, we generate six stochastic realisations of each field in the batch, and pass the ensemble mean of the realisations to the content loss. This approach averages over generated fine scale features and only applies the content loss on the patterns which are expected to be consistent across realisations. Both approaches allow generated variability at small scales but encourage convergence at large scales.

We experimented with two different content loss functions: mean absolute error (MAE), and cumulative rank probability score (CRPS). MAE is commonly used as a content loss in deep learning, but is a deterministic measure. CRPS is defined as 

\begin{align}
    CRPS(F, x) &= \int_{-\infty}^{\infty}\Big(F(y)- H(y - x)\Big)^2dy
\end{align}
where $F$ represents the CDF of the predicted distribution, $H$ is the Heaviside function, and $x$ is the ground truth value. The CRPS metric returns lower values to distributions whose mass is centred around the ground truth value. In a deterministic setting, the PDF of $y$ is a delta functtion, and the CRPS generalises to MAE, so is appropriate to use as a content loss. To apply the CRPS, we calculated an empirical CRPS metric for each pixel, using the ground truth value and values from the stochastic sampling realisations, and then took the mean across pixels.

Throughout this paper, we will use the following naming conventions to specify models: \[\text{Training}^{\text{Content Loss}}_{\text{Noise Level}},\] where Training is represented respectively by $F$ or $S$ for frequency separation and stochastic sampling, and noise level is represented by $NC$ for noise covariate, and $low$, $medium$, and $full$ for noise injection. For example, $S^{CRPS}_{full}$ specifies a model using stochastic sampling, the CRPS content loss, and full noise injection. Note that we have not investigated all combinations of Generator parameters, as some only make sense in combination or in specific settings. 

\subsection{Validation}
Quality assessment in image generation problems often poses a challenge, because there are multiple, often competing, metrics that could be used. Potential metric priorities include convergence in realization (pixelwise or at large scales), or convergence in statistical features such as spatial covariance or pixelwise marginal distributions. In general, we will consider a combination of these factors, depending on the problem. As noted earlier, while deterministic pixel-wise error metrics are important, they should not be relied on too much due to the double-penalty problem. Following from \citet{harris2022generative}, \citet{annau2023algorithmic} and \citet{ravuri2021skilful}, we used a Radially Averaged Spectral Power metric (RASP) for comparing spatial variance of different scales (or alternatively, the covariance structure) between the generated fields and the ground truth. We calculated RASP by first performing a 2-dimensional Fourier transform on each field, averaging the amplitudes within annular rings centred at wavenumber zero,  and then averaging power densities across at least 1000 fields. Ideally, spectral power at each spatial scale should be the same in the generated and truth fields; to aid in visual comparison, we standardised amplitudes at each wavenumber by the amplitude of the ground truth field in the corresponding bin. A value below one represents less spectral power at the given wavenumber in the generated field than in the ground truth, and a value above one suggests more spectral power than in the ground truth field. This quantity allows assessment of biases across spatial scales.

To assess the quality of stochastic realisations for a given conditional distribution, we compared pixel-wise marginal distributions (where possible), and used rank histograms. For the synthetic data, we estimated the true marginal distributions using Kernel Density Estimates of 500 realisations sampled from the truth distributions, and  compared these pixelwise to the equivalent marginal distributions of 500 stochastic realisations from the trained model. To test pixelwise convergence across the whole domain, we calculated the empirical Kolmogorov-Smirnov statistic (KS) at each pixel and investigated the distribution of these KS statistics for a given model. The KS statistic is defined as \[D = \sup_x |G(x) - T(x)|\] where $G$ and $T$ represent the CDFs of the generated and true distributions, respectively.  

We calculated rank histograms by generating 96 stochastic realisations of HR fields for each of 50 randomly selected LR conditioning fields, and determined the pixelwise rank of the truth field in the ensemble of generated fields. That is, for each pixel, we calculated 
\begin{equation}
    k_n = rank(x, \langle g_1,...,g_{96} \rangle)
\end{equation}
where $x$ is the truth value for the pixel, and $g$ are the ensemble forecast members. We then used CDFs of histograms to investigate the distribution of the ranks. If the model is well calibrated, the truth field should be indistinguishable from any ensemble member, and so the rank histogram should estimate a uniform distribution, corresponding to a linear CDF. Conversely, if the CDF has more weight at the tails, corresponding to a U-shaped histogram, then the model is underdispersive (most truth points fall outside the range of generated realisations).

The rank histogram described above is computed using the rank of each pixel. However, since calibration of extremes is often most important, we used a modified rank histogram to asses performance with regards to spatial extremes. Here, we randomly selected 400 samples from the test set, generated 100 stochastic realisations of each, and then calculated the 0.999 and 0.001 quantiles across the 16384 pixels in each field. We then produced a rank histogram of true quantiles compared to the 96 generated quantiles, for each sample: 
\begin{equation}
    m_n = rank(q(x),\langle q(g_1),...,q(g_{96}) \rangle )
\end{equation}
where $q$ represents the quantile over the field.

\section{Results}
In this section, we investigate results pertaining to two main classes of distributions: the distribution of HR fields conditioned on LR covariates, and full HR fields across samples. Consideration of the conditional distribution, $p(HR | LR)$ allows investigation of the stochastic calibration: given a set of LR covariates, what is the distribution of the HR fields? The full distribution, 
\begin{equation}
    p(HR) = \int p(HR | LR) p(LR) ~ dLR \approx \frac{1}{n}\sum_{k \in LR} p(HR | LR_k)
\end{equation} 
represents the full distribution across LR conditioning sets, where $n$ is the number of conditioning sets being considered. Throughout the results, we focus on four models: $F_{nc}^{MAE}$ (the baseline; partial frequency separation with noise covariates and MAE content loss), $F_{full}^{MAE}$ (partial frequency separation with full noise injection and MAE content loss), $S_{full}^{MAE}$ (stochastic sampling with full noise injection and MAE content loss), and $S_{full}^{CRPS}$ (stochastic sampling with full noise injection and CRPS content loss). Note that since CRPS is a probabilistic loss, it cannot be applied to the $F$ class models. 

\subsection{Synthetic Data}
\subsubsection{Noise Injection}
\begin{figure}[H]
\noindent\includegraphics[width=\textwidth,angle=0]{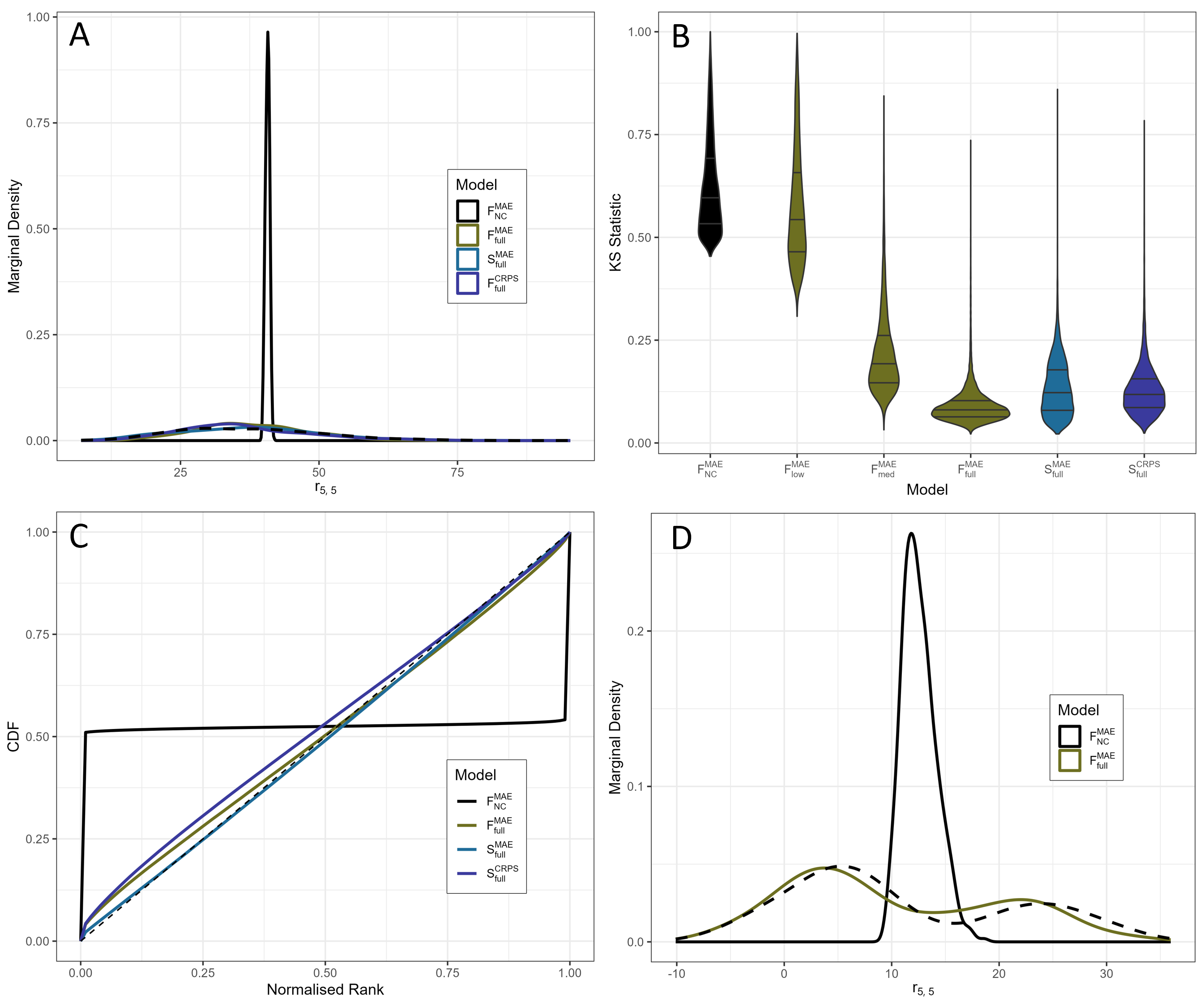}
\caption{a) Kernel density estimates (KDEs) of marginal distributions of $p(HR|LR)$ for the unimodal synthetic dataset for one example pixel (i = 5, j = 5) for the true distribution and generated distributions. KDEs are based on 500 realisations for a single conditioning field for each distribution. Dashed line shows true marginal distribution. b) Violin plot showing KS statistic values comparing generated marginal conditional distributions to ground truth distributions for all pixels. Statistics are calculated for each pixel individually, using 500 realisations of a single conditioning field. Lines show 0.25, 0.5, and 0.75 quantiles, respectively. c) CDF of rank histogram on unimodal synthetic data, with four models, showing calibration of conditional distributions. Dashed line shows reference uniform distribution. Rank histograms were calculated across 100 randomly selected conditioning fields, with 96 HR realisations generated for each. d) KDEs of marginal conditional distributions for one example pixel of a bimodal dataset, comparing true (dashed line) and generated distributions. Distributions were estimated using the same approach as in a).}
\label{fig:fourpanel}
\end{figure}

Models with full noise injection performed substantially better overall than the baseline models at matching the true marginal distribution. A representative example of pixelwise marginal distributions  for the truth and generated fields is shown in figure \ref{fig:fourpanel}a. The baseline $F_{NC}^{MAE}$ model produced highly underdispersive distributions, while noise injection models were able to match the true distributions well. This result held across all pixels:  KS statistics comparing the true marginal distributions with those from noise injection models were significantly smaller than with the baseline model (figure \ref{fig:fourpanel}b). Both $S$ class models had slightly larger median KS statistics than $F_{full}^{MAE}$ , but still showed good distributional matching. 

Decreasing the number of noise injection layers in the Generator decreased performance of conditional distributions (figure \ref{fig:fourpanel}b). Low noise injection ($F_{low}^{MAE}$) showed slight improvement over the baseline model, but still produced underdispersive results. Medium noise injection ($F_{med}^{MAE}$) had improved statistics, but was still generally underdispersive, and full noise injection ($F_{full}^{MAE}$) created results with the best agreement of pixelwise marginal distributions.

Rank histograms provide another tool for investigating calibration of conditional distributions, averaged across multiple conditioning fields. Rank histograms showed good calibration for all models with full noise injection, and severe underdispersion for the baseline model (figure \ref{fig:fourpanel}c). The $S_{full}^{MAE}$ model showed almost perfect calibration, but all noise injection models performed well. 

Learning multimodal distributions is challenging for GANs; they tend to show mode collapse, in which  distributions are collapsed to the conditional mean \citep{saatci2017bayesian}. We found that using the bimodal dataset (equation 10), the $F_{full}^{MAE}$ model could learn both modes of the marginal distributions. The baseline models usually showed mode collapse and could not meaningfully recreate any of the marginal distributions (figure \ref{fig:fourpanel}d). 

\begin{figure}[H]
  \noindent\includegraphics[width=\textwidth,angle=0]{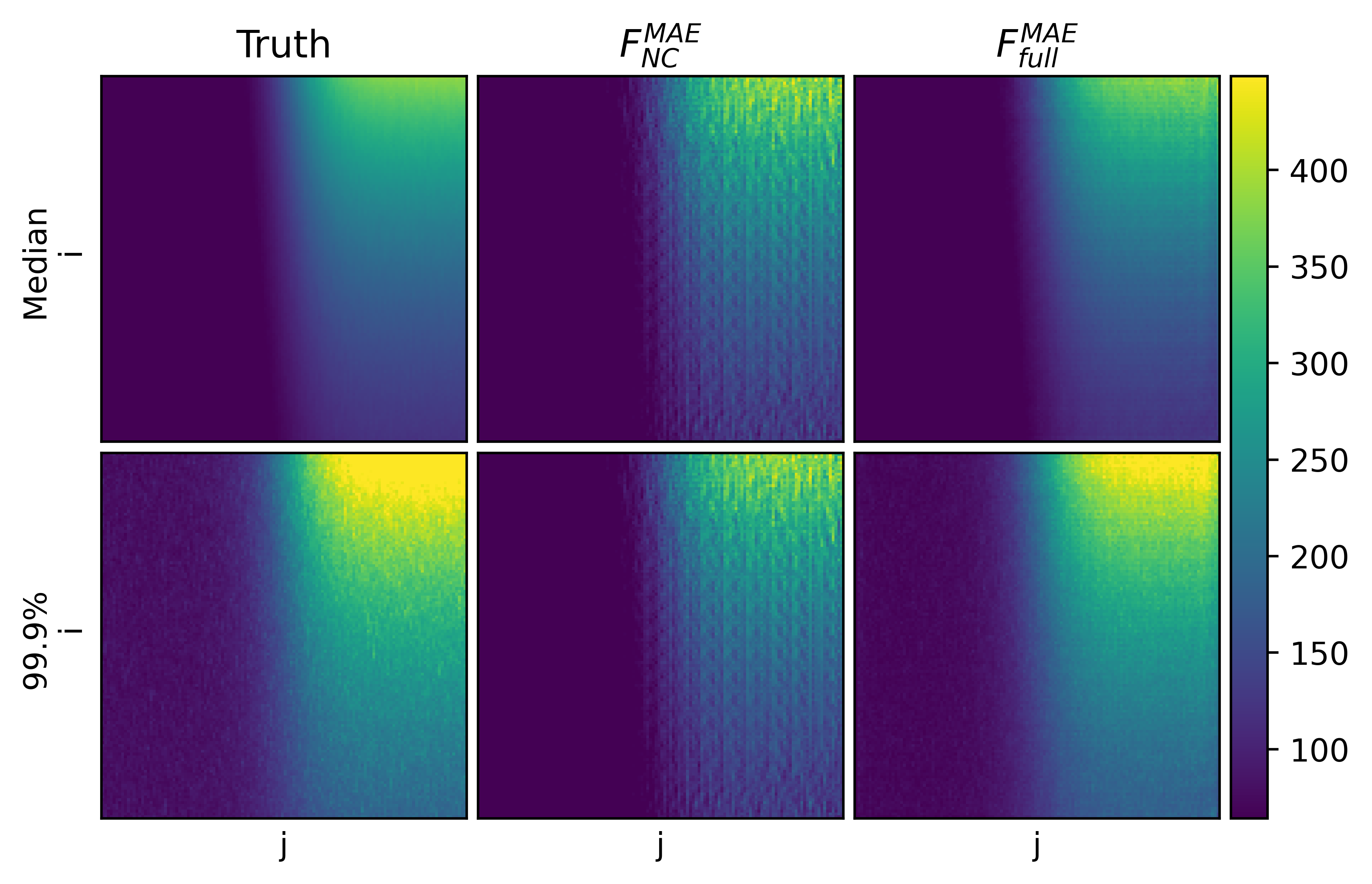}
  \caption{Spatial fields of median and 99.9 percentiles of the full distributions across samples for ground truth, and generated data from two models, using the unimodal synthetic dataset (equation 3). }\label{f3}
\end{figure}

Investigating results of the full distribution, $p(HR)$, statistics of generated fields had fewer artifacts and were closer to the ground truth statistics using the $F_{full}^{MAE}$ model than the baseline model (figure \ref{f3}). Even after training metrics had converged, the baseline model showed noticeable traces of the convolutional filters as checkerboard artifacts, which were not apparent in the $F_{full}^{MAE}$ model. The baseline model also substantially underestimated the 99.9 percentiles, especially at the highest values. These were better captured (although not perfectly) by the $F_{full}^{MAE}$ model.

\begin{figure}[H]
  \noindent\includegraphics[width=\textwidth,angle=0]{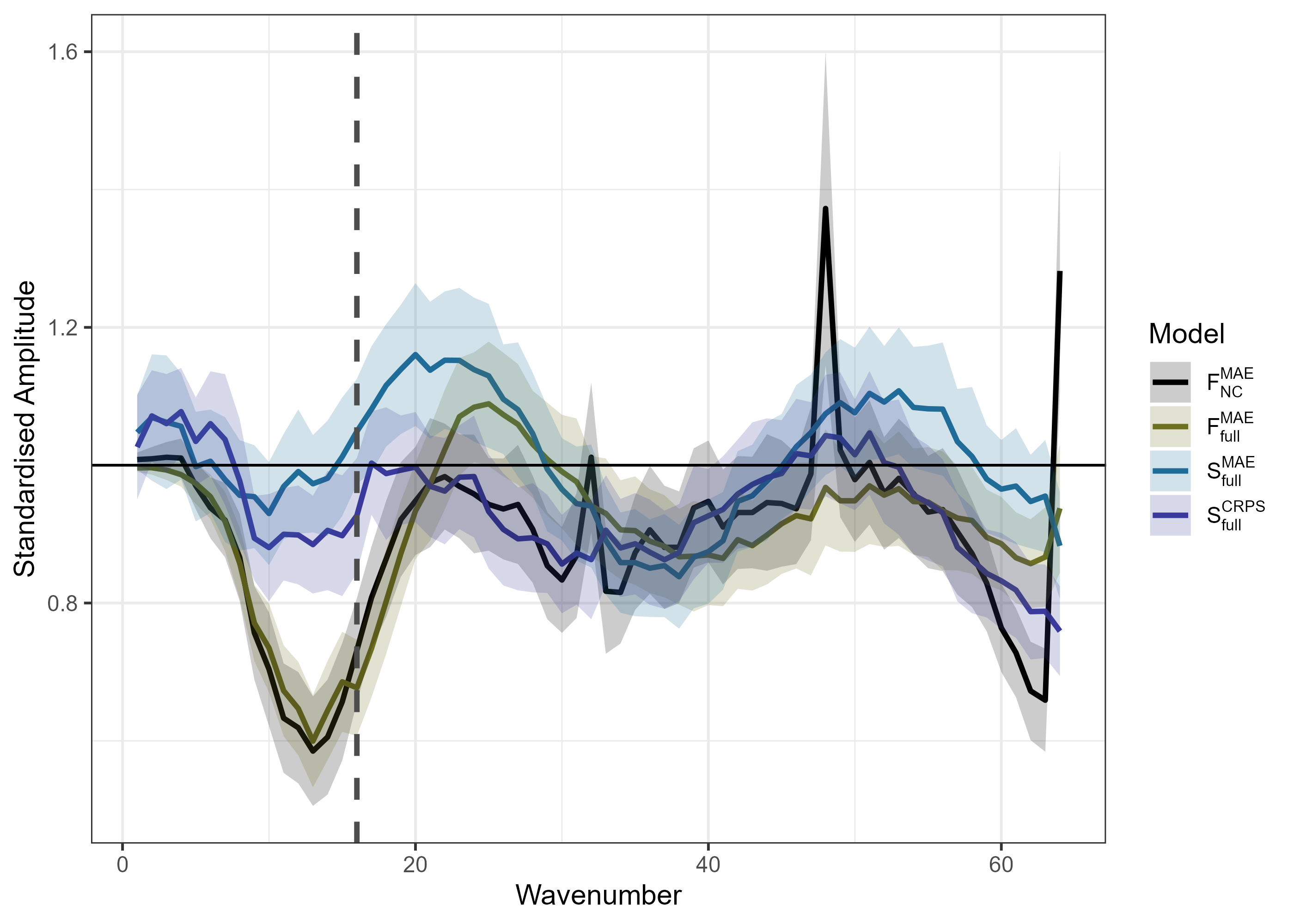}\\
  \caption{Radially averaged spectral power (RASP) for four models. Values are standardised to amplitudes of ground truth wavenumbers, so perfectly matched spectral power occurs at one. Solid lines and shaded regions respectively show mean and +/- one standard deviations across 1200 randomly selected samples. Dashed line indicates wavenumber corresponding to LR pixel size. }\label{f4}
\end{figure}

As well as better capturing pixelwise marginal variability, all noise injection models performed better at representing covariance patterns. The RASP metrics (figure \ref{f4}) demonstrate that the baseline model showed too little power for a range of low wavenumbers, and then spurious spikes at other wavenumbers. The $F_{full}^{MAE}$ did not show the spikes, but still had a lower power bias at low wavenumbers. Both $S$ class models substantially improved the representation of power at low wavenumbers, and in general were the closest to the true spectral power accross all wavenumbers. There was no obvious difference in spectral power between the two $S$ class models using this metric.

Overall, synthetic data experiments showed that models using full noise injection were substantially better at capturing conditional distributions than the baseline model. In general, all models with noise injection performed comparably well. There was also noticeable improvement in the quality of the full distributions using noise injection, and the $S$ class models showed improved ability to represent spatial dependence. 

\subsection{Wind Downscaling Case Study}
As above, we first present results for conditional distributions, before moving to the full distributions across time. Note that while all models predicted both zonal (eastward) and meridional (northward) wind components, results were generally similar, and unless otherwise stated, we only show results for meridional components. 

\begin{figure}[H]
  \noindent\includegraphics[width=\textwidth,angle=0]{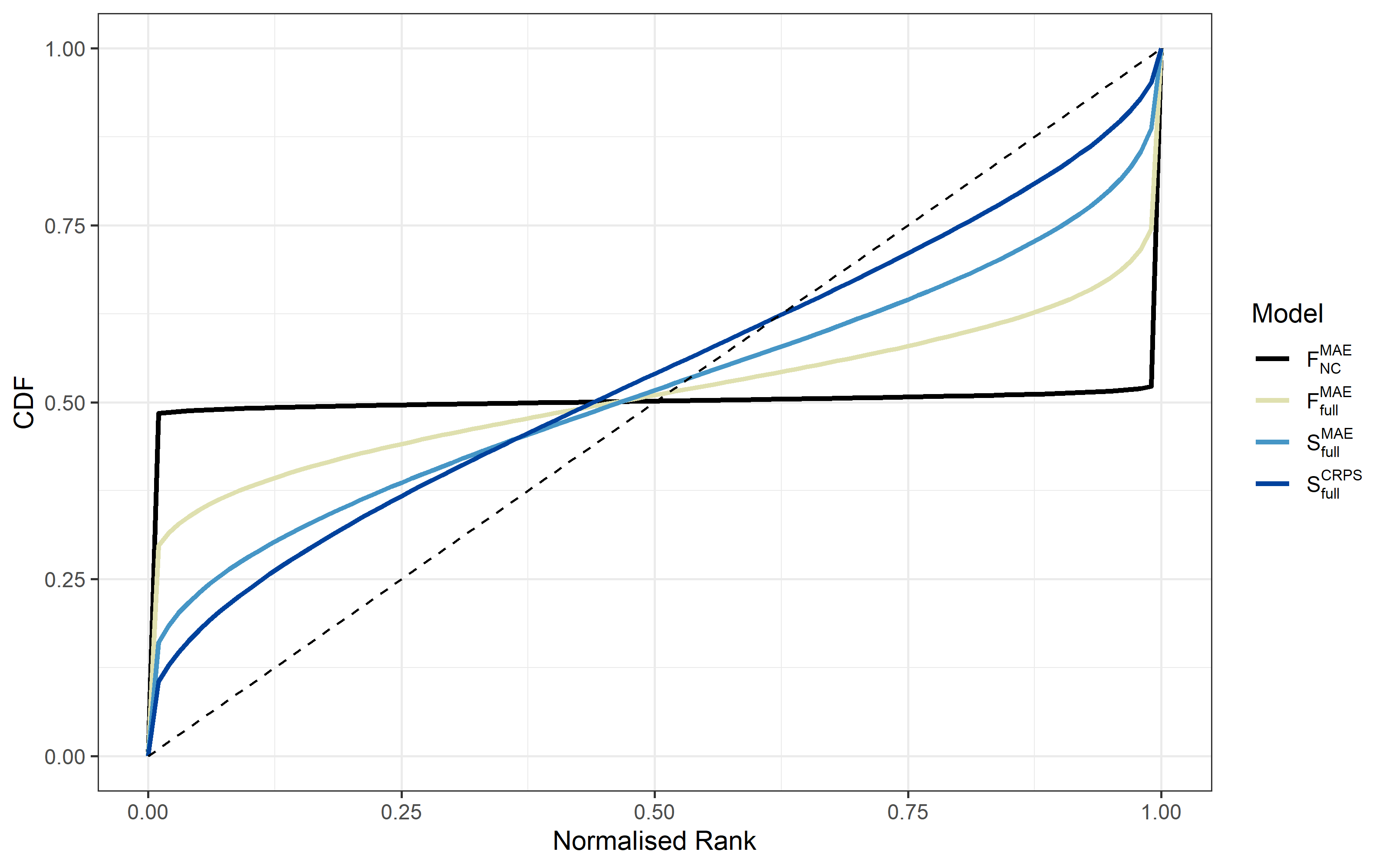}\\
  \caption{CDFs of rank histograms for meridional wind components, using four different models. Rank histograms were calculated across 100 randomly selected conditioning fields, with 96 HR realisations Generator for each. Dashed line shows reference uniform CDF.}\label{f6}
\end{figure}

Even with noise injection, the $F$ class models applied to a realistic downscaling problem produced underdispersive results (figure \ref{f6}).  While the $F^{MAE}_{full}$ model showed substantial improvement over the baseline model, it did not fully capture the conditional variability. Both $S^{MAE}_{full}$ and $S^{CRPS}_{full}$ models had better calibration than the $F$ class models; the $S^{CRPS}_{full}$ model, while still showing some underdispersion, performed best among the models considered (figure \ref{f6}).  

\begin{figure}[H]
  \noindent\includegraphics[width=\textwidth,angle=0]{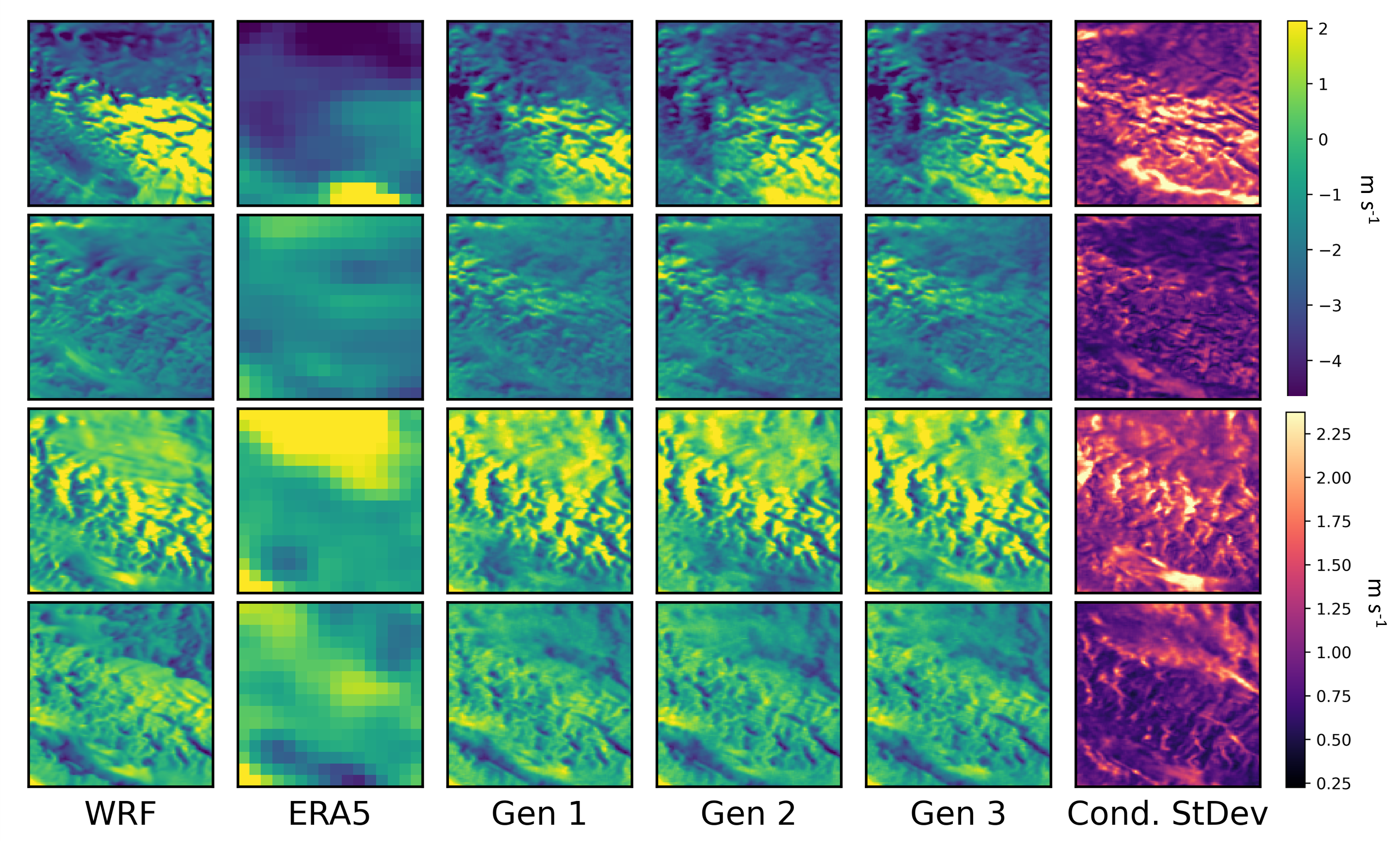}\\
  \caption{Example meridional (top two rows) and zonal (bottom two rows) wind fields for coastal BC using the $S_{full}^{CRPS}$ model. Rows correspond to four randomly selected timesteps in the test data, and columns show, from left to right, WRF (i.e. ground truth), ERA5 (input conditioning field),  three generated realisations, and the conditional standard deviations across 500 realisations.}\label{f7}
\end{figure}

With the $S_{full}^{CRPS}$ model, individual realisations were visually realistic, and showed noticeable differences in spatial patterns given a single set of conditioning fields (figure \ref{f7}). Examples of standard deviation fields showed that the conditional distribution varies substantially with the state of the conditioning fields.

\begin{figure}[H]
  \noindent\includegraphics[width=\textwidth,angle=0]{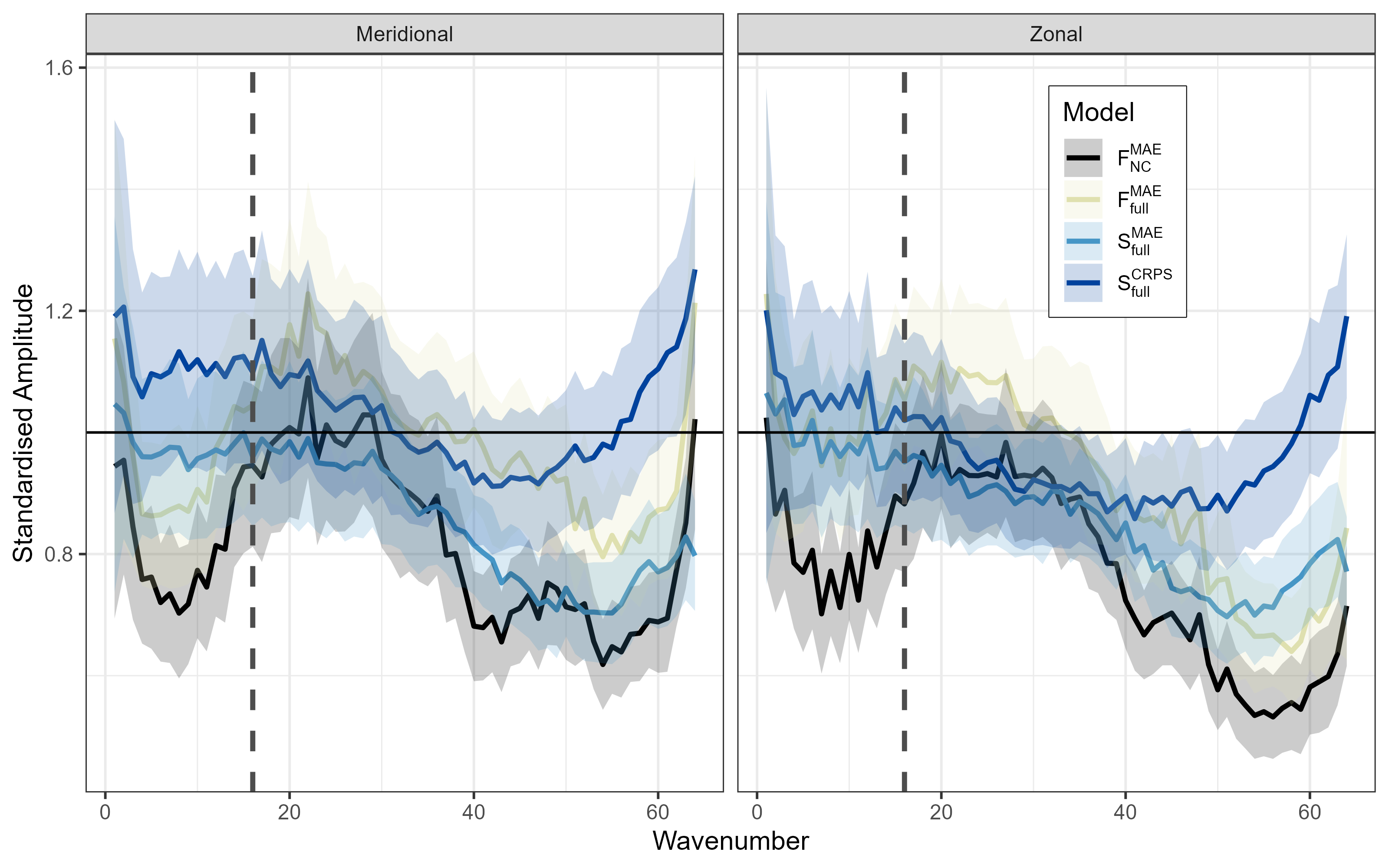}\\
  \caption{RASP metric (mean +/- 1 SD) standardised to ground truth values for zonal and meridional wind fields. Spectral powers are calculated across 1200 randomly selected fields. Dashed line shows wavenumber corresponding to LR grid size.}\label{f8}
\end{figure}

Looking at the full distribution, spectral power was also better calibrated with the $S$ class models (figure \ref{f8}). The baseline model produced low spectral power at both low and high wavenumbers; performance of the $F_{full}^{MAE}$ model was better but followed similar patterns and showed a substantial high bias at intermediate wavenumbers. The $S_{full}^{CRPS}$ model generally had the most similar distribution of spectral power to the truth fields, although it showed a modest high power bias at low and high wavenumbers, especially for meridional wind components. All models using the $MAE$ loss function showed low-power bias at high wavenumbers, consistent with the suppression of fine-scale features by this deterministic metric.  

\begin{figure}[H]
  \noindent\includegraphics[width=\textwidth]{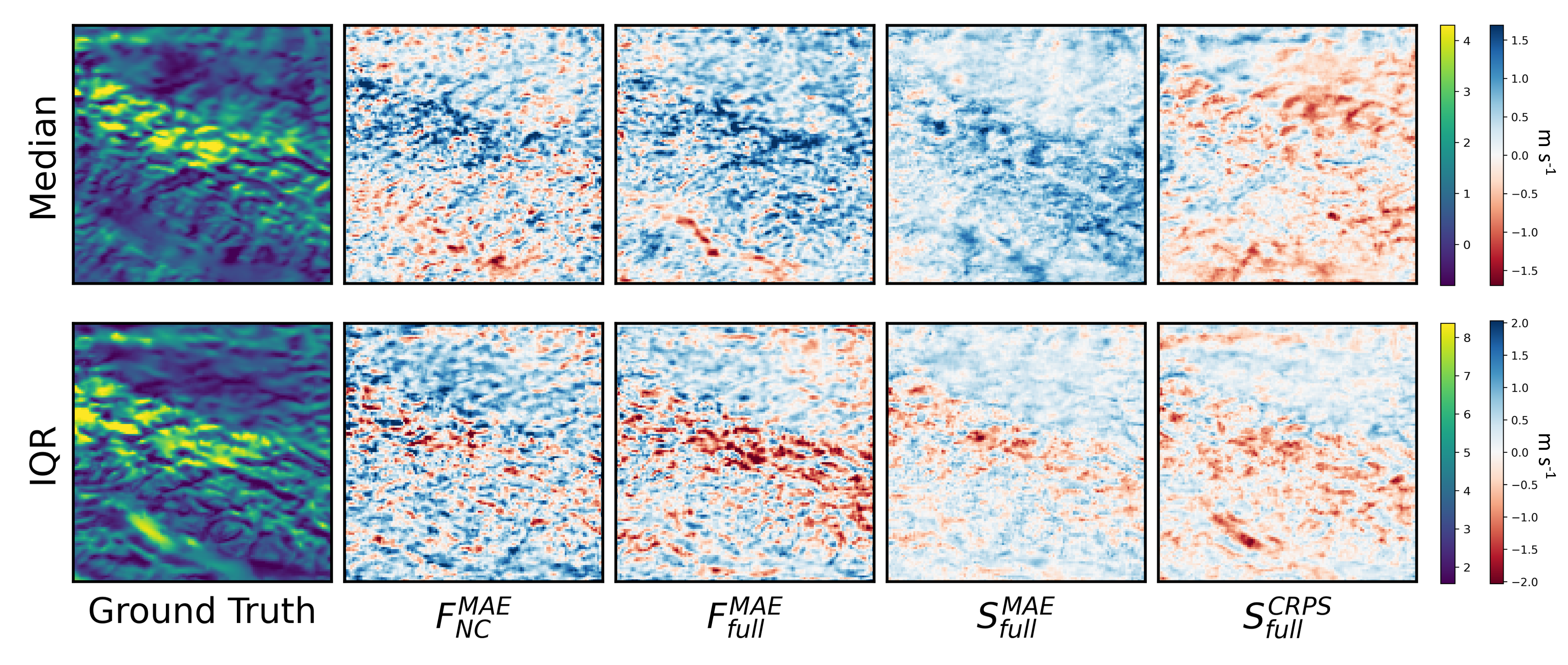}\\
  \caption{Median and inter-quartile range (IQR) of full distribution of the test dataset for meridional wind fields. The first column shows truth statistics, followed by difference fields for each of the four models (truth - model).}\label{fbulkstats}
\end{figure}

All models produced visually realistic downscaled HR fields, and pixelwise marginal statistics of full distributions generally matched the true statistics well  (figure \ref{fbulkstats}). Differences in median and inter-quartile range were spatially smoother with the $S$ class models, suggesting these models were able to capture fine spatial patterns better.  The baseline model substantially underestimated inter-quartile range (IQR) in many locations. This bias was improved with the $F^{MAE}_{full}$ model, although it overestimated IQR in certain areas, and the $S$ class models further improved the results. 

\begin{figure}[H]
\noindent\includegraphics[width=\textwidth]{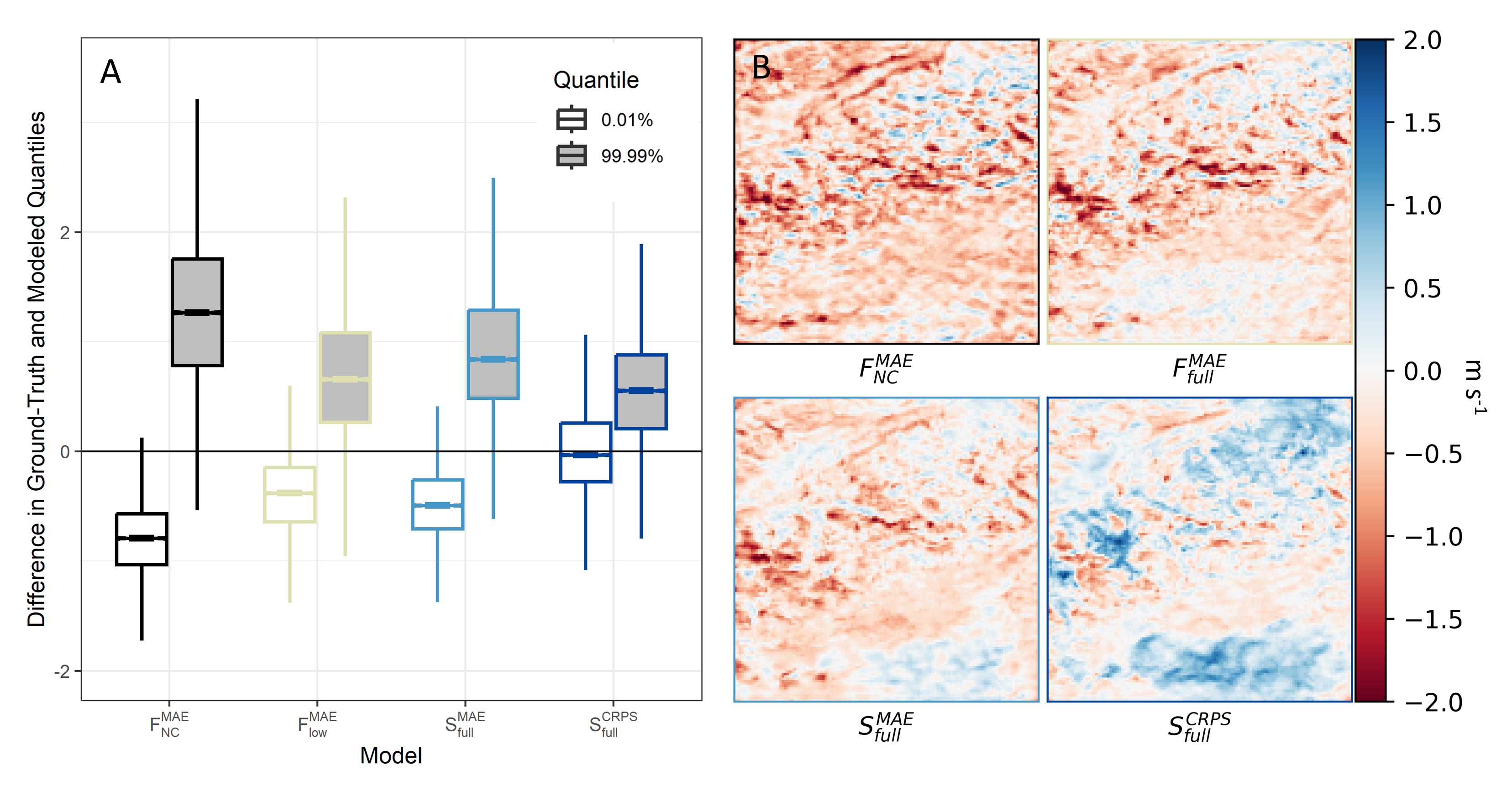}\\
\caption{Calibration of moderate extremes for meridional wind fields over full distributions. a) Boxplots of distributions of difference in 99.99 and 0.01 percentiles of ground truth and generated realisations for four models, based on 500 realisations for each of 350 randomly selected conditioning fields. Values below zero represent model overestimation; values above zero represent underestimation. b) Difference maps of 0.01 percentiles of ground truth and generated realisations for four models.}
\label{fig:extreme}
\end{figure}

Investigating the tail behaviour of the full distributions, we found that models that were better at learning conditional distributions performed better at predicting accurate extremes (figure \ref{fig:extreme}). Comparing moderately large marginal extremes (99.99 and 0.01 percentiles) across pixels, the $S_{full}^{CRPS}$ model had the least biased estimate of extremes while biases from the $F_{nc}^{MAE}$ were largest, underestimating 99.99 percentiles and overestimating 0.01 percentiles (figure \ref{fig:extreme}a). This pattern is also apparent in difference maps of the extremes (cf. 0.01 percentile in figure \ref{fig:extreme}b). The map for $F_{nc}^{MAE}$ takes the same sign almost everywhere, whereas the map for $S_{full}^{CRPS}$ shows reduced systematic bias and is fairly well centred around zero for the 0.01 percentiles. 

\begin{figure}[H]
  \noindent\includegraphics[width=\textwidth,angle=0]{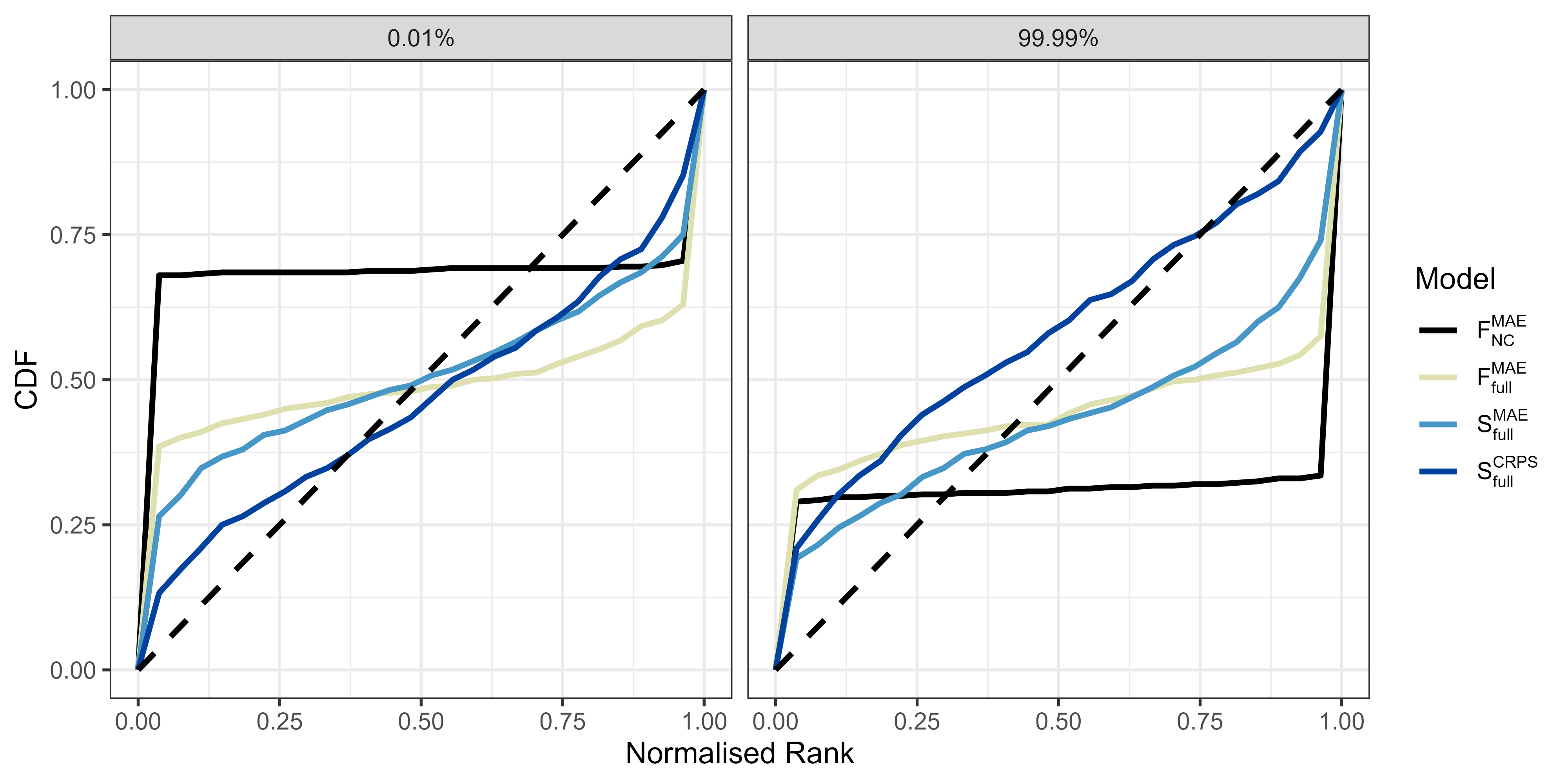}
  \caption{CDFs of rank histograms based on 0.01 and 99.99 percentiles of meridional wind fields over 400 conditioning fields, with 96 realisations of each field. Dashed lines represent CDF of a uniform distribution.}\label{f11}
\end{figure}

Calibration of spatial extremes was also improved in the $S$ class models (figure \ref{f11}). Here, rank histograms represent stochastic calibration of models in regards to spatial extreme values (i.e., large or small quantiles of values across the domain). Rank histograms of 0.01 and 99.99 percentiles across wind fields showed that the $S_{full}^{CRPS}$ model was the least underdispersive, and had less bias than the other models (figure \ref{f11}). In contrast, the rank histogram of the baseline $F_{NC}^{MAE}$ model showed the model systematically overpredicted 0.01 percentiles, and underpredicted 99.99 percentiles. 

\begin{figure}[H]
\noindent\includegraphics[width=\textwidth,angle=0]{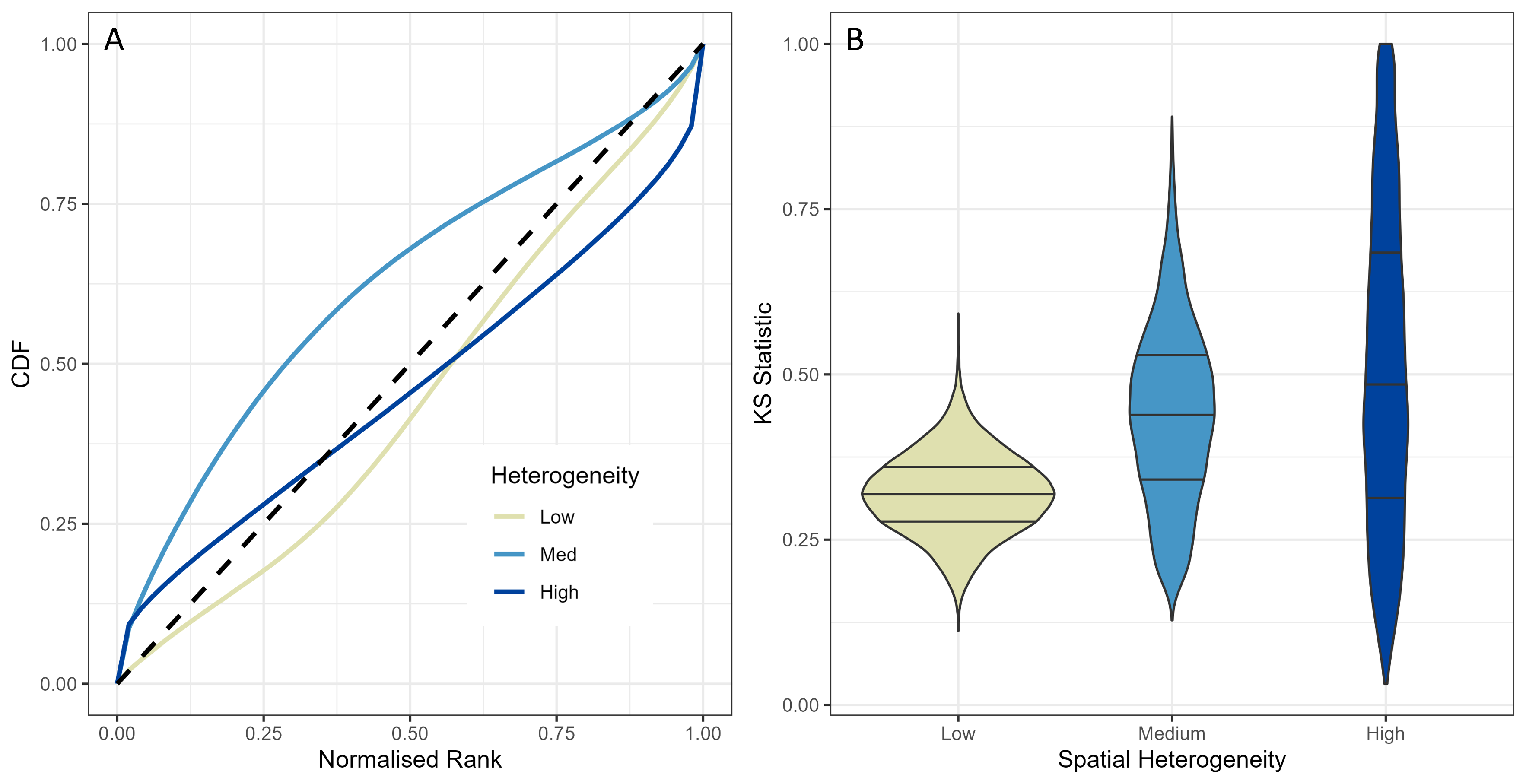}

\caption{Comparison of stochastic calibration of the $F_{full}^{MAE}$ model based on synthetic data with low, moderate and high spatial heterogeneity. a) CDFs of rank histograms based on all pixels of 50 random conditiong fields with 96 realisations of each. b) Distribution of pixel-wise KS statistics between generated and true marginal distributions, using 500 stochastic realisations for a single conditioning field. }
\label{fig:spat_hetero}
\end{figure}

To investigate why the $F_{full}^{MAE}$ model showed excellent calibration on the synthetic dataset, but was underdispersive when used to generate high-resolution wind fields, we tested the effect of increased spatial heterogeneity on synthetic data models (cf. Section 2a). Models trained on datasets with high heterogeneity showed a greater degree of underdispersion than those with low or moderate heterogeneity (figure \ref{fig:spat_hetero}). Investigating the KS statistics of the pixelwise marginal distributions showed that distributions from datasets with high heterogeneity had very large ranges in quality, whereas those from low heterogeneity data were more consistent and better matches on average (figure \ref{fig:spat_hetero}b). The rank histograms of the results from these datasets showed similar patterns - the low heterogeneity model was relatively well calibrated, and the high heterogeneity model was underdispersive. Interestingly, while the moderate level model did not show as much underdispersion, it showed the most bias, more often than not overestimating values.  These results suggest that the $F$ class models struggle to produce good stochastic calibrations when there is a high degree of spatial heterogeneity in the fields - as is often the case in a realistic setting. 

\newpage
\section{Discussion and Conclusions}

This paper discusses three main classes of GAN-based downscaling models distinguished by: noise type (noise covariate vs noise injection), training method (frequency separation vs stochastic sampling) and content loss type (deterministic MAE vs probabilistic CRPS). We aim to improve stochastic calibration, creating models that can successfully sample from the full range of the conditional distribution (i.e., for a given large-scale atmospheric state, we want the model to generate the full range of local weather possibilities). We first present a novel network architecture, where many layers of noise at different resolutions are injected into the Generator. Compared to the baseline $F_{NC}^{MAE}$ model, our architecture performs better at capturing the conditional variability of the data (thus reducing underdispersion), and achieves good calibration on synthetic data. We then introduce a stochastic training method which greatly improves stochastic calibration and spatial structure, especially when combined with the probabilistic CRPS metric in the $S_{full}^{CRPS}$ model. The $S_{full}^{CRPS}$ model shows improved skill at estimating marginal conditional distributions, as well as marginal and spatial statistics of the full distribution. 

Most conditional GANs in super resolution are deterministic, and recent attempts at reintroducing stochasticity have added noise fields as additional covariates \citep[e.g.,][]{leinonen2020stochastic}. Our approach of injecting noise directly into the convolutional layers fundamentally differs in that it adds noise to latent representations deep inside the network, instead of to the input. When noise is introduced as a covariate at the beginning of the network, we hypothesise that the network will learn low weights for the noise layers in order to optimise the loss function. By adding noise to the latent representations, we slightly alter features inside the network, leading to better representation of conditional and full distributions. Our approach is similar to the nESRGAN+ architecture \citep{nesrganplus} which injects noise inside the Residual in Residual Dense Blocks, but we inject noise one level deeper, inside the Dense Blocks, to alter the output of the basic convolutional layers. In contrast to the nESRGAN+, we also use noise injection at both the low and high resolution, allowing for more scales of stochasticity. Interestingly, we never experienced problems with overdisperison as we increased the number of noise injection layers; marginal distributions were closest with the maximum possible number of noise injection layers for a given architecture. 

Stochastic sampling is an approach adapted from \citet{harris2022generative}, in which the content loss function is calculated on a set of stochastic realisations. While the $F_{full}^{MAE}$ model showed excellent calibration on the synthetic data and much improved performance than the $F_{NC}^{MAE}$ model, it still failed to fully capture the variability in the wind downscaling application. The $S$ class models, especially in combination with the CRPS loss function, resulted in substantially improved performance. This improvement could be the result of a few different factors. First, by using multiple realisations of each field in the loss function, the network has more information to use for backpropagation. Second, CRPS is a stochastic metric which aims to quantify distributional matching; it seems reasonable that it thus improves the stochastic calibration. In contrast to our study, \citet{harris2022generative}  did not find an improvement with using CRPS in the loss function. This difference in results could be due to differences in Generator architecture. As \citet{harris2022generative} did not use noise injection, perhaps the generated stochasticity was less suitable to optimisation with the CRPS metric. 

All models considered in this study use a separation of spatial scales in the content loss in an effort to address the double penalty problem inherent in the ill-conditioned nature of climate downscaling. In our synthetic data experiments, we found that partial frequency separation (PFS), as described in \citet{annau2023algorithmic}, resulted in well calibrated output. However, this method did not perform as well in the realistic setting of wind component downscaling, motivating consideration of the stochastic sampling approach. Fundamentally, PFS and stochastic sampling have similar goals: allowing the adversarial loss freedom to create small-scale features, while rewarding consistency between generated and conditioning fields at large scales. While PFS achieves this by only sending low frequency information to the content loss,  the stochastic sampling approach applies the content loss function across an ensemble of stochastic realisations, thus ``smoothing out" the smaller scale features of the generated fields. The stochastic sampling approach is likely more accurate than PFS - instead of arbitrarily choosing a frequency for separation, the sample conditional means define the transition from conditioning scales to sampling scales. Indeed, we found that for downscaling wind components, stochastic sampling always outperformed models using PFS. A practical challenge with stochastic sampling is that it uses more computational resources during training than PFS models, as each of the stochastic realisations have to be used during backpropagation. Most notably, stochastic sampling nearly doubled the amount of memory required during training, and it increased training time by about 50\% (using a stochastic batch size of six). In practice, the choice of training approach will likely depend on the desired outcome and computational resources available. While the stochastic sampling models performed substantially better at capturing conditional and full distributions, they only performed slightly better at capturing the spatial patterns of single conditioning fields. Thus, if the goal is to produce downscaling without needing to capture conditional variability, it may be prudent to use PFS and reduce training requirements. It is of course possible that the improvement gained from using stochastic sampling on capturing the conditional distributions will depend strongly on the fields being considered. 

Wind fields show a high level of spatial heterogeneity, which we expect is responsible for the difficulties experienced by the $F_{full}^{MAE}$ model in capturing the conditional variability. Our experiment with spatial heterogeneity showed that even with synthetic data, increasing heterogeneity lead to increased underdispersion and bias in conditional means. High heterogeneity will generally lead pixel-wise metrics to be more sensitive - slight shifts in features from the truth fields will tend to result in poor pixel-wise metrics compared to similar shifts in fields with low heterogeneity. Future work could consider wind fields and other pertinent physical fields across areas with different degrees of spatial heterogeneity.

From a theoretical standpoint, stochastic downscaling is an appealing approach as it provides a way to quantify the range of solutions to the underdetermined problem of climate downscaling. In addition, we have found that improved distributional estimates lead to better representation of extreme events, both spatially and temporally. A model which is able to accurately sample from a distribution will sometimes draw samples from the tails of the distribution, whereas models with substantial underdispersion will tend to only sample from the conditional mean. \citet{harris2022generative} found that their models were underdispersive when applied to extremes; it would be interesting to see if the improvements made here could improve their analysis of precipitation downscaling.

Modelling extreme events is of utmost importance to climate adaptation, and these events are often more challenging to model than averages \citep{thompson2013means}. Infrastructure needs to handle precipitation and wind extremes; most heat-related human health issues occur during extreme heat waves \citep{kephart2022city}. Generally, statistical downscaling has not been successful at capturing extreme events, and while dynamical downscaling can perform better, it is too computationally intensive for some practical applications. Our study has shown that by improving the ability of GANs to make distributional estimates, we are able to obtain better estimates of extremes, both spatially and temporally, often with a marginal increase in computational cost. Hence, deep-learning based downscaling shows promise as a statistical downscaling strategy with the ability to more accurately capture extremes. Further research will be required to determine whether these results generalise to a non-stationary climate (e.g., across time periods). If so, deep-learning downscaling could become an essential part of climate adaptation for estimating future extremes.

\clearpage
\acknowledgments
We thank Nicolaas Annau for providing code and for valuable discussion. We also acknowledge Dr. Colin Mahony and Dr. Karen Price for helpful comments and reviews. Funding for this work was provided by the British Columbia Ministry of Forests through the ClimatEx program, and by the British Columbia Graduate Fellowship.

%
%
\datastatement
Our fully stochastic GAN with CRPS loss is implemented with PyTorch in https://github.com/nannau/ClimatExML/tree/stochastic. LR conditioning fields are available from ERA5 (https://cds.climate.copernicus.eu/) and HR WRF training data are available at https://www.gwfnet.net/Metadata/Record/T-2020-05-28-Q1KtfEjVdz0aBmauuvG9r9w.

\bibliographystyle{ametsocV6}
\bibliography{main}

\begin{thebibliography}{25}
\providecommand{\natexlab}[1]{#1}
\providecommand{\url}[1]{\texttt{#1}}
\renewcommand{\UrlFont}{\rmfamily}
\providecommand{\urlprefix}{URL }
\expandafter\ifx\csname urlstyle\endcsname\relax
  \providecommand{\doi}[1]{https://doi.org/\discretionary{}{}{}#1}\else
  \providecommand{\doi}{https://doi.org/\discretionary{}{}{}\begingroup \urlstyle{rm}\Url}\fi
\providecommand{\eprint}[2][]{\url{#2}}

\bibitem[{Afshari et~al.(2023)Afshari, Vogel,, and Chockalingam}]{afshari2023statistical}
Afshari, A., J.~Vogel, and G.~Chockalingam, 2023: Statistical downscaling of {SEVIRI} land surface temperature to {WRF} near-surface air temperature using a deep learning model. \textit{Remote Sensing}, \textbf{15~(18)}, 4447.

\bibitem[{Annau et~al.(2023)Annau, Cannon,, and Monahan}]{annau2023algorithmic}
Annau, N.~J., A.~J. Cannon, and A.~H. Monahan, 2023: Algorithmic hallucinations of near-surface winds: Statistical downscaling with generative adversarial networks to convection-permitting scales. \textit{Artificial Intelligence for the Earth Systems}, \textbf{2~(4)}, e230\,015.

\bibitem[{Arjovsky et~al.(2017)Arjovsky, Chintala,, and Bottou}]{arjovsky2017wasserstein}
Arjovsky, M., S.~Chintala, and L.~Bottou, 2017: Wasserstein generative adversarial networks. \textit{International conference on machine learning}, PMLR, 214--223.

\bibitem[{Chen et~al.(2014)Chen, Liu, Liu,, and Li}]{chen2014spatial}
Chen, F., Y.~Liu, Q.~Liu, and X.~Li, 2014: Spatial downscaling of {TRMM 3B43} precipitation considering spatial heterogeneity. \textit{International Journal of Remote Sensing}, \textbf{35~(9)}, 3074--3093.

\bibitem[{Fischer et~al.(2021)Fischer, Sippel,, and Knutti}]{fischer2021increasing}
Fischer, E., S.~Sippel, and R.~Knutti, 2021: Increasing probability of record-shattering climate extremes. \textit{Nature Climate Change}, \textbf{11~(8)}, 689--695.

\bibitem[{Goodfellow et~al.(2014)Goodfellow, Pouget-Abadie, Mirza, Xu, Warde-Farley, Ozair, Courville,, and Bengio}]{goodfellow2014generative}
Goodfellow, I.~J., J.~Pouget-Abadie, M.~Mirza, B.~Xu, D.~Warde-Farley, S.~Ozair, A.~Courville, and Y.~Bengio, 2014: Generative adversarial networks. \textit{arXiv preprint arXiv:1406.2661}.

\bibitem[{Harris et~al.(2022)Harris, McRae, Chantry, Dueben,, and Palmer}]{harris2022generative}
Harris, L., A.~T. McRae, M.~Chantry, P.~D. Dueben, and T.~N. Palmer, 2022: A generative deep learning approach to stochastic downscaling of precipitation forecasts. \textit{arXiv preprint arXiv:2204.02028}.

\bibitem[{Jiang et~al.(2020)Jiang, Huang,, and Hu}]{jiang2020single}
Jiang, Z., Y.~Huang, and L.~Hu, 2020: Single image super-resolution: Depthwise separable convolution super-resolution generative adversarial network. \textit{Applied Sciences}, \textbf{10~(1)}, 375.

\bibitem[{Kephart et~al.(2022)}]{kephart2022city}
Kephart, J.~L., and Coauthors, 2022: City-level impact of extreme temperatures and mortality in {L}atin {A}merica. \textit{Nature Medicine}, \textbf{28~(8)}, 1700--1705.

\bibitem[{Ledig et~al.(2017)}]{ledig2017photo}
Ledig, C., and Coauthors, 2017: Photo-realistic single image super-resolution using a generative adversarial network. \textit{Proceedings of the IEEE conference on computer vision and pattern recognition}, 4681--4690.

\bibitem[{Leinonen et~al.(2020)Leinonen, Nerini,, and Berne}]{leinonen2020stochastic}
Leinonen, J., D.~Nerini, and A.~Berne, 2020: Stochastic super-resolution for downscaling time-evolving atmospheric fields with a generative adversarial network. \textit{IEEE Transactions on Geoscience and Remote Sensing}, \textbf{59~(9)}, 7211--7223.

\bibitem[{Li et~al.(2019)Li, Li, Zhang, Chen, Kurkute, Scaff,, and Pan}]{li2019high}
Li, Y., Z.~Li, Z.~Zhang, L.~Chen, S.~Kurkute, L.~Scaff, and X.~Pan, 2019: High-resolution regional climate modeling and projection over western canada using a weather research forecasting model with a pseudo-global warming approach. \textit{Hydrology and Earth System Sciences}, \textbf{23~(11)}, 4635--4659.

\bibitem[{Lucas-Picher et~al.(2021)Lucas-Picher, Arg{\"u}eso, Brisson, Tramblay, Berg, Lemonsu, Kotlarski,, and Caillaud}]{lucas2021convection}
Lucas-Picher, P., D.~Arg{\"u}eso, E.~Brisson, Y.~Tramblay, P.~Berg, A.~Lemonsu, S.~Kotlarski, and C.~Caillaud, 2021: Convection-permitting modeling with regional climate models: Latest developments and next steps. \textit{Wiley Interdisciplinary Reviews: Climate Change}, \textbf{12~(6)}, e731.

\bibitem[{MacKenzie and Mahony(2021)MacKenzie, and Mahony}]{mackenzie2021ecological}
MacKenzie, W.~H., and C.~R. Mahony, 2021: An ecological approach to climate change-informed tree species selection for reforestation. \textit{Forest Ecology and Management}, \textbf{481}, 118\,705.

\bibitem[{Maraun(2013)}]{maraun2013bias}
Maraun, D., 2013: Bias correction, quantile mapping, and downscaling: Revisiting the inflation issue. \textit{Journal of Climate}, \textbf{26~(6)}, 2137--2143.

\bibitem[{Mirza and Osindero(2014)Mirza, and Osindero}]{mirza2014conditional}
Mirza, M., and S.~Osindero, 2014: Conditional generative adversarial nets. \textit{arXiv preprint arXiv:1411.1784}.

\bibitem[{Price and Rasp(2022)Price, and Rasp}]{price2022increasing}
Price, I., and S.~Rasp, 2022: Increasing the accuracy and resolution of precipitation forecasts using deep generative models. \textit{International Conference on Artificial Intelligence and Statistics}, PMLR, 10\,555--10\,571.

\bibitem[{{Rakotonirina} and {Rasoanaivo}(2020){Rakotonirina}, and {Rasoanaivo}}]{nesrganplus}
{Rakotonirina}, N.~C., and A.~{Rasoanaivo}, 2020: Esrgan+ : Further improving enhanced super-resolution generative adversarial network. \textit{ICASSP 2020 - 2020 IEEE International Conference on Acoustics, Speech and Signal Processing (ICASSP)}, 3637--3641.

\bibitem[{Ravuri et~al.(2021)}]{ravuri2021skilful}
Ravuri, S., and Coauthors, 2021: Skilful precipitation nowcasting using deep generative models of radar. \textit{Nature}, \textbf{597~(7878)}, 672--677.

\bibitem[{Saatci and Wilson(2017)Saatci, and Wilson}]{saatci2017bayesian}
Saatci, Y., and A.~G. Wilson, 2017: Bayesian gan. \textit{Advances in neural information processing systems}, \textbf{30}.

\bibitem[{Skamarock et~al.(2001)Skamarock, Klemp,, and Dudhia}]{skamarock2001prototypes}
Skamarock, W.~C., J.~B. Klemp, and J.~Dudhia, 2001: Prototypes for the {WRF} (weather research and forecasting) model. \textit{Preprints, Ninth Conf. Mesoscale Processes, J11--J15, Amer. Meteorol. Soc., Fort Lauderdale, FL}, Vol.~1.

\bibitem[{Thompson et~al.(2013)Thompson, Beardall, Beringer, Grace,, and Sardina}]{thompson2013means}
Thompson, R.~M., J.~Beardall, J.~Beringer, M.~Grace, and P.~Sardina, 2013: Means and extremes: building variability into community-level climate change experiments. \textit{Ecology Letters}, \textbf{16~(6)}, 799--806.

\bibitem[{Wang et~al.(2020)Wang, Chen, Yang, Bi,, and Yu}]{wang2020state}
Wang, L., W.~Chen, W.~Yang, F.~Bi, and F.~R. Yu, 2020: A state-of-the-art review on image synthesis with generative adversarial networks. \textit{IEEE Access}, \textbf{8}, 63\,514--63\,537.

\bibitem[{Wang et~al.(2018)Wang, Yu, Wu, Gu, Liu, Dong, Qiao,, and Change~Loy}]{wang2018esrgan}
Wang, X., K.~Yu, S.~Wu, J.~Gu, Y.~Liu, C.~Dong, Y.~Qiao, and C.~Change~Loy, 2018: Esrgan: Enhanced super-resolution generative adversarial networks. \textit{Proceedings of the European conference on computer vision (ECCV) workshops}, 0--0.

\bibitem[{Wilks(2010)}]{wilks2010use}
Wilks, D.~S., 2010: Use of stochastic weathergenerators for precipitation downscaling. \textit{Wiley Interdisciplinary Reviews: Climate Change}, \textbf{1~(6)}, 898--907.

\end{thebibliography}

\end{document}